\documentclass[conference]{IEEEtran}
\IEEEoverridecommandlockouts
\usepackage{cite}
\usepackage{amsmath,amssymb,amsfonts}
\usepackage{algorithmic}
\usepackage{graphicx}
\usepackage{textcomp}
\usepackage{xcolor}
\usepackage[numbers, sort&compress]{natbib}
\usepackage{dsfont}
\usepackage[utf8]{inputenc} 
\usepackage[T1]{fontenc}    
\usepackage{url}            
\usepackage{booktabs}       
\usepackage{amsfonts}       
\usepackage{nicefrac}       
\usepackage{microtype}      

\usepackage{listings}
\usepackage{caption}
\usepackage{bbm}
\usepackage{arydshln}

\usepackage{placeins}
\usepackage{nicefrac}
\usepackage{xurl}
\usepackage{enumitem}
\usepackage{tabularx}
\usepackage{nicefrac}
\usepackage{makecell}
\usepackage{xspace}
\usepackage{graphbox}
\usepackage{pifont}
\usepackage{enumitem}
\usepackage{colortbl}
\usepackage{multirow}
\usepackage{diagbox}
\usepackage{array}
\usepackage{wrapfig}
\usepackage{subcaption}
\usepackage{footmisc}
\usepackage{color}
\usepackage{stfloats}
\usepackage{slashbox}
\usepackage{diagbox}
\usepackage{float}
\usepackage{mathrsfs}
\usepackage{apptools}
\usepackage{bbm}
\usepackage{soul}

\graphicspath{ {figures} }

\usepackage[dvipsnames]{xcolor}
\usepackage{tcolorbox}
\usepackage{url}

\definecolor{iccvblue}{rgb}{0.21,0.49,0.74}
\usepackage[pagebackref,breaklinks,colorlinks,citecolor=iccvblue, linkcolor=red]{hyperref}

\definecolor{codeblue}{rgb}{0.13, 0.29, 0.53}
\definecolor{codegreen}{rgb}{0, 0.6, 0}
\definecolor{codegray}{rgb}{0.5, 0.5, 0.5}
\definecolor{codepurple}{rgb}{0.58, 0, 0.82}
\definecolor{backcolour}{rgb}{0.95, 0.95, 0.92}

\lstdefinestyle{pythonstyle}{
    backgroundcolor=\color{backcolour},   
    commentstyle=\color{codegreen},
    keywordstyle=\color{codepurple},
    numberstyle=\tiny\color{codegray},
    stringstyle=\color{codeblue},
    basicstyle=\ttfamily\footnotesize,
    breakatwhitespace=false,         
    breaklines=true,                 
    captionpos=b,                    
    keepspaces=true,                 
    numbers=left,                    
    numbersep=5pt,                  
    showspaces=false,                
    showstringspaces=false,
    showtabs=false,                  
    tabsize=4
}

\def\BibTeX{{\rm B\kern-.05em{\sc i\kern-.025em b}\kern-.08em
    T\kern-.1667em\lower.7ex\hbox{E}\kern-.125emX}}
\begin{document}

\title{Video Evidence to Reasoning: Efficient Video Understanding via Explicit Evidence Grounding}

\author{Yanxiang Huang, Guohua Gao, Zhaoyang Wei \\Department of Applied Mathematics, The Hong Kong Polytechnic University}

\maketitle

\begin{abstract}
Large Vision-Language Models (LVLMs) face a fundamental dilemma in video reasoning: they are caught between the prohibitive computational costs of verbose reasoning and the hallucination risks of efficient, ungrounded approaches. To resolve this, we introduce the Chain of Evidence (CoE), a novel framework that architecturally decouples and co-optimizes perceptual grounding and reasoning efficiency. CoE incorporates two core innovations: (1) A lightweight Evidence Grounding Module (EGM) that acts as a query-guided filter, dynamically identifying and extracting a compact set of high-fidelity visual evidence; and (2) An Evidence-Anchoring Protocol optimized via Reinforcement Learning. Crucially, we design a composite reward mechanism that enforces process alignment, compelling the model to strictly reference identified temporal anchors during deduction, thereby mitigating hallucinations. To enable this, we construct CoE-Instruct, a large-scale dataset (164k samples) featuring a novel dual-annotation schema for separate perception and reasoning supervision. Extensive experiments on five benchmarks, including Video-MME, MVBench, and VSI-Bench, demonstrate that CoE-enhanced models establish a new state-of-the-art. They significantly outperform existing methods in accuracy, proving CoE to be a powerful and practical paradigm for reliable video understanding.
\end{abstract}

\begin{IEEEkeywords}
Spatio-temporal  Grounding, Video Understanding, Evidence Reasoning
\end{IEEEkeywords}

\section{Introduction}
\label{sec:intro}

Large Vision-Language Models (LVLMs) have demonstrated remarkable capabilities in multimodal understanding, extending the step-by-step reasoning abilities of Large Language Models (LLMs)~\citep{wei2022chain, kojima2022large} to the visual domain. This has led to state-of-the-art systems, from proprietary models like GPT-4o~\citep{achiam2023gpt} to open-source giants like InternVL~\citep{chen2024expanding}, that can process and reason about complex visual scenes. However, when applied to video understanding, these models confront a fundamental challenge: the sheer volume of temporal data exposes a critical trade-off between the \textbf{accuracy of perceptual grounding} and the \textbf{efficiency of the reasoning process}.

This trade-off manifests in two primary failure modes. On one hand, to enhance reasoning accuracy, models are often prompted to generate verbose Chain-of-Thought (CoT) explanations~\citep{wang2024videocot, wu2024vstar}. While beneficial, this approach leads to excessive token consumption and high inference latency, rendering it impractical for real-world video applications. On the other hand, approaches aimed at improving efficiency often resort to complex, multi-stage pipelines involving auxiliary networks for key-frame selection~\citep{han2024videoespresso, hu2025mllm} or the generation of intermediate representations like scene graphs~\citep{hao2024video}. While these methods reduce the visual input, they often introduce significant inference overhead and do not address the verbosity of the subsequent reasoning phase. Crucially, both paradigms fail to address a core issue: the reasoning process is not explicitly and reliably anchored to the visual evidence, leading to factual inconsistencies, or "hallucinations"~\cite{alayrac2022flamingo}. Existing methods either lack explicit temporal grounding in their reasoning traces or require costly and complex data generation procedures~\citep{wang2024videocot, han2024videoespresso}, limiting their scalability and generalizability.

\begin{figure}[t]
    \centering
    \includegraphics[width=1\linewidth]{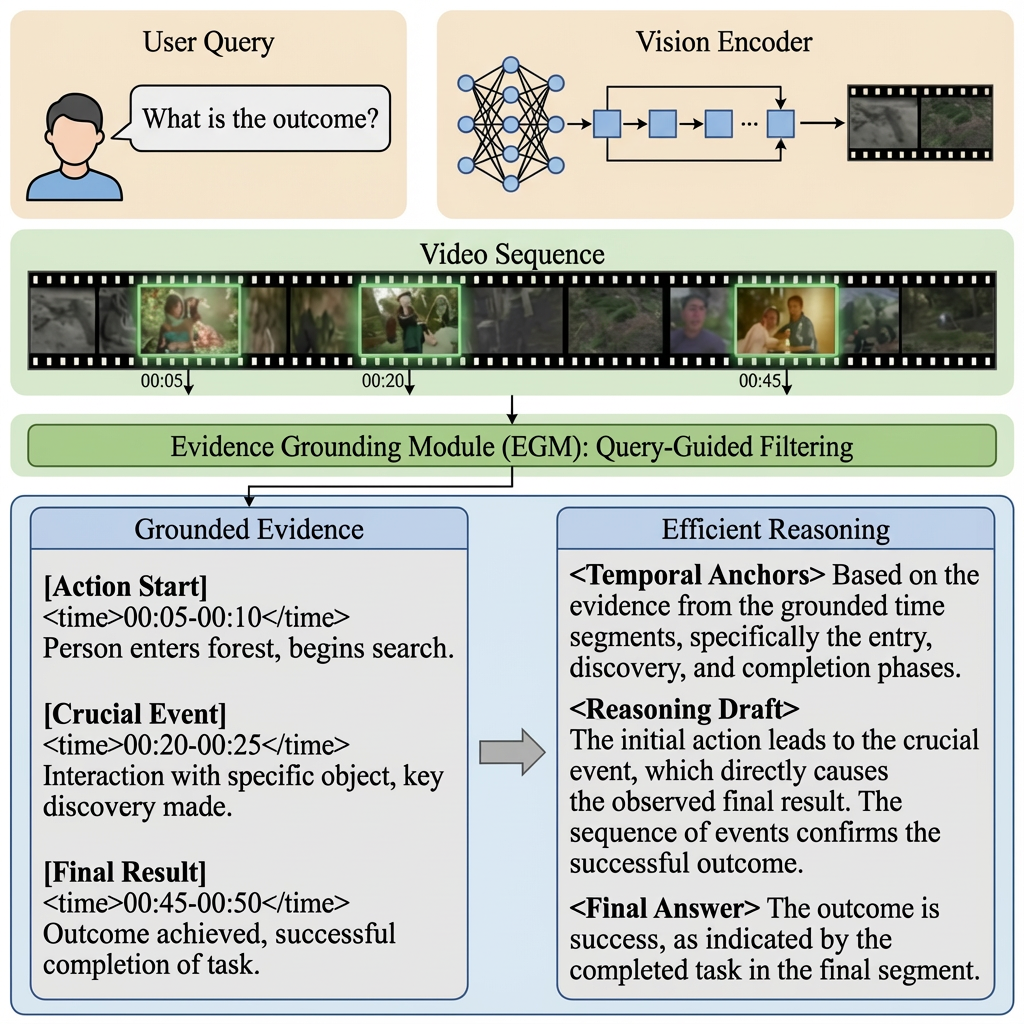} 
    \caption{The framework of CoE, where the EGM first extracts compact visual evidence based on the query, enabling the LLM to perform efficient and grounded reasoning.}
    \label{fig:framework}
\vspace{-5pt}
\end{figure}

To remedy these limitations, we introduce the \textbf{Chain of Evidence (CoE)}, a novel and unified framework designed to fundamentally resolve the tension between perceptual accuracy and reasoning efficiency in video LVLMs. CoE is built on the principle of \textbf{decoupling perception from reasoning}: it trains the model to first explicitly identify and extract a structured set of spatio-temporal visual evidence, and then perform concise, logical reasoning conditioned upon this high-fidelity evidence set. As illustrated in Fig.~\ref{fig:framework}, our approach enables the model to generate a reasoning process that is both grounded in specific video frames and computationally efficient.
Our contributions are threefold:
\begin{enumerate}[leftmargin=*, itemsep=2pt, topsep=2pt]
    \item We propose the \textbf{Chain of Evidence (CoE) framework}, a new paradigm for video reasoning that, for the first time, architecturally decoules and co-optimizes perceptual grounding and reasoning efficiency within a single, end-to-end model.
    
    \item We construct \textbf{CoE-Instruct}, a large-scale (164k samples) training dataset with a novel dual-annotation schema that provides separate, explicit supervision for both evidence grounding and efficient reasoning tasks.
    
    \item Through extensive experiments on five challenging video understanding benchmarks, including \textbf{VIDEO-MME}~\citep{fu2024video}, \textbf{MVBENCH}~\citep{li2023mvbench}, and \textbf{VSI-BENCH}~\citep{yang2024think}, we demonstrate that our CoE-enhanced models establish a new state-of-the-art, significantly outperforming existing methods in accuracy while achieving substantial reductions in token usage and inference latency.
\end{enumerate}

\section{The Chain of Evidence Framework}
\label{sec:method}
Existing video LVLMs often face a dilemma: pursuing rigor through verbose reasoning incurs prohibitive costs, while prioritizing efficiency leads to hallucinated details. To resolve this, we introduce the \textbf{Chain of Evidence (CoE)}, a framework that architecturally decouples and co-optimizes perceptual grounding and reasoning. As shown in Fig.~\ref{fig:framework}, CoE operates in two phases: \textit{Perception} via an Evidence Grounding Module (EGM), and \textit{Deduction} via a structured CoE Reasoning Protocol.

\subsection{Architectural Design: A Hierarchical Approach}
\label{sec:architecture}

\begin{figure}[t]
    \centering
    \includegraphics[width=1\linewidth]{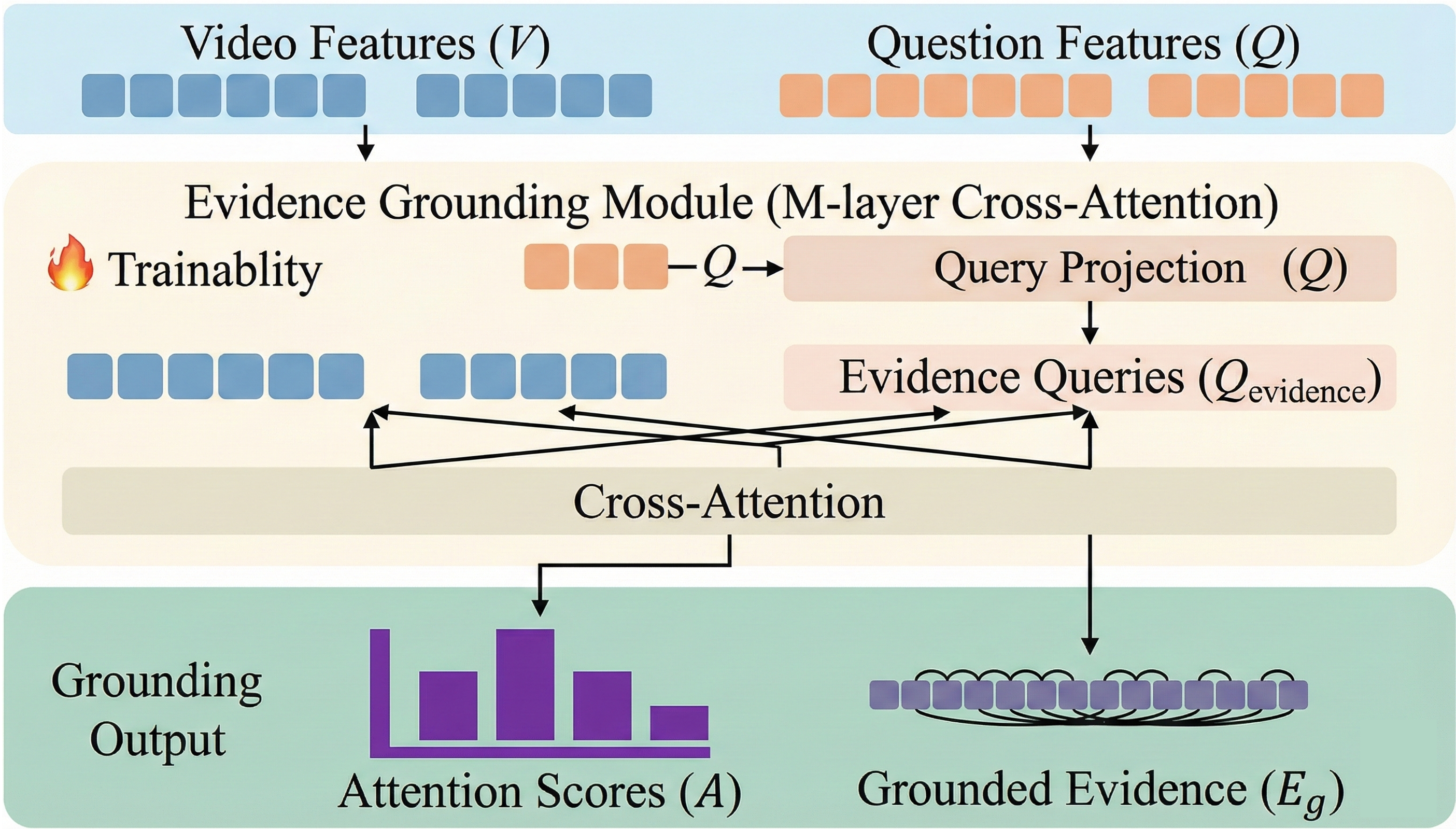} 
    \caption{The EGM acts as a query-guided filter between the vision encoder and the language decoder, enabling a hierarchical processing of visual information.}
    \label{fig:EGM}
\vspace{-5pt}
\end{figure}

Our CoE framework is built upon a standard LVLM backbone, comprising a Vision Transformer (ViT) encoder and a large language model (LLM) decoder. Its core innovation lies in the introduction of a lightweight \textbf{Evidence Grounding Module (EGM)}, positioned between the encoder and decoder, as illustrated in Fig.~\ref{fig:EGM}. The ViT encoder first processes the video into a sequence of frame features $V = \{v_1, v_2, \dots, v_N\} \in \mathbb{R}^{N \times D_v}$. The EGM then dynamically filters these features based on the user's query, and finally, the LLM receives the refined evidence to generate the output.

\paragraph{The Evidence Grounding Module (EGM)}
The EGM is designed as a \textbf{query-guided visual information filter}, mimicking the human cognitive process of searching for clues with a specific question in mind. It takes as input the sequence of video frame features $V$ and the embedded question features $Q \in \mathbb{R}^{L \times D_l}$. We implement the EGM as a shallow, $M$-layer cross-attention network. Within each layer, we first project the question features $Q$ to generate a set of learnable \textbf{evidence query vectors}, $Q_{\text{evidence}} \in \mathbb{R}^{K \times D_v}$, where $K$ is a hyperparameter for the maximum number of evidence pieces. These queries then attend to the video features:
\begin{align}
\footnotesize
    A &= \text{softmax}\left(\frac{Q_{\text{evidence}} V^T}{\sqrt{D_v}}\right) \in \mathbb{R}^{K \times N} \label{eq:attention} \\
    E_g &= A \cdot V \in \mathbb{R}^{K \times D_v} \label{eq:grounded_features}
\end{align}
The resulting matrix $A$ is the \textbf{attention weight matrix}, providing a differentiable mechanism to identify key frames. $E_g$ is the final output, termed \textbf{Grounded Evidence Features}, a condensed feature sequence of length $K$ where each vector is a query-weighted aggregation of all frame features. By transforming the lengthy video sequence $V$ into a short, query-relevant sequence $E_g$, the EGM fundamentally addresses the computational burden on the subsequent LLM.

\subsubsection{The CoE Reasoning Protocol}
Unlike standard Chain-of-Thought (CoT) which relies on implicit visual memories, CoE enforces a \textbf{white-box reasoning process}. Upon receiving the condensed evidence $E_g$, the LLM does not immediately generate an answer. Instead, it follows a strict \textbf{Evidence-Anchoring Protocol} composed of three explicit steps:

\begin{enumerate}[leftmargin=*, noitemsep, topsep=0pt]
    \item \textbf{Explicit Anchoring:} The model first translates the visual features $E_g$ into interpretable time ranges (e.g., \texttt{<00:05-00:10>}), explicitly stating \textit{where} it is looking.
    \item \textbf{Evidence-Interleaved Deduction:} The model generates a \texttt{Reasoning\_Draft} that must strictly reference the identified anchors to build logical links (e.g., \textit{"At [00:05], the action starts..."}).
    \item \textbf{Conclusion:} Finally, the model derives the \texttt{Answer} based solely on the verified draft.
\end{enumerate}

To implement this, we design a structured system prompt that differentiates CoE from standard CoT. A comparison of the expected output format is shown below:

\vspace{5pt}
\noindent\fbox{
\parbox{0.95\linewidth}{
\footnotesize
\textbf{Standard CoT:} ``The video shows a man running... Therefore, the answer is Yes.'' (Implicit grounding, prone to hallucination) \\
\textbf{Our CoE Protocol:} \\
\texttt{<Temporal Anchors>} 00:05-00:10; 00:45-00:50 \texttt{</Temporal Anchors>} \\
\texttt{<Reasoning Draft>} Based on the entry at 00:05 and the result at 00:45... \texttt{</Reasoning Draft>} \\
\texttt{<Answer>} Yes \texttt{</Answer>}
}
}
\vspace{5pt}

This protocol ensures that every generated conclusion is traceable to specific visual evidence ($E_g$), fundamentally transforming reasoning from a black-box generation to an evidence-driven deduction.

\subsection{Decoupled Joint Training Strategy for SFT}
\label{sec:training}

To train the EGM to accurately identify evidence and the LLM to effectively reason upon it, we devise a decoupled multi-task training strategy. This strategy relies on our specially curated \textbf{CoE-Instruct} dataset, where each sample contains a \texttt{Question}, a list of \texttt{Key\_Frame\_Indices}, a concise \texttt{Reasoning\_Draft}, and a final \texttt{Answer}. Our total loss function, $L_{\text{CoE}}$, is composed of two distinct components that supervise perception and reasoning separately:
\begin{equation}
    L_{\text{CoE}} = L_{\text{grounding}} + \lambda \cdot L_{\text{reasoning}}
    \label{eq:total_loss}
\end{equation}
where $\lambda$ is a hyperparameter balancing the two tasks.

\paragraph{Grounding Loss ($L_{\text{grounding}}$)}
This loss is designed exclusively to supervise the EGM. We first derive a \textbf{frame importance score vector}, $a_{\text{scores}} \in \mathbb{R}^{N}$, by applying max-pooling over the query dimension ($K$) of the attention weight matrix $A$ from Eq.~\ref{eq:attention}:
\begin{equation}
\footnotesize
    a_{\text{scores}}[i] = \max_{j=1 \dots K} A[j, i] \quad \text{for } i=1 \dots N
\end{equation}
Simultaneously, we convert the ground-truth \texttt{Key\_Frame\_Indices} into a binary target vector $y_{\text{target}} \in \{0, 1\}^N$. The grounding loss is then the \textbf{Binary Cross-Entropy (BCE)} between the predicted scores and the target vector:
\begin{align}
    L_{\text{grounding}} &= \text{BCE}(a_{\text{scores}}, y_{\text{target}}) \nonumber \\
    &= -\frac{1}{N} \sum_{i=1}^{N} \left[ y_i \log(\sigma(a_i)) + (1-y_i) \log(1-\sigma(a_i)) \right]
    \label{eq:bce_loss}
\vspace{-5pt}
\end{align}
where $\sigma$ is the sigmoid function. This loss directly encourages the EGM to assign high attention to key frames and low attention to irrelevant ones.

\paragraph{Reasoning Loss ($L_{\text{reasoning}}$)}
To teach the \textit{CoE Protocol}, we train the LLM to generate the full sequence of $Y = [\texttt{Anchors}; \texttt{Draft}; \texttt{Answer}]$ conditioned on $E_g$. This compels the model to learn the dependency between evidence and logic:
\begin{equation}
\footnotesize
    L_{\text{reasoning}} = -\sum_{t} \log p(y_t | y_{<t}, E_g, Q)
\end{equation}

\subsection{Reinforcement Learning from Evidence Feedback}
\label{sec:rl_feedback}

To refine the SFT-tuned model, we employ a reinforcement learning strategy that rewards both the accuracy of the final answer and the quality of the intermediate evidence chain. This encourages the model to not only be correct, but to be correct for the right reasons.

\paragraph{Reward Modeling}
To strictly enforce the \textit{Evidence-Interleaved Deduction}, we augment the reward function with process supervision to bridge the gap between perception and reasoning:
\begin{equation}
\footnotesize
    R(x, y) = w_g \cdot \text{F1}(A_{\text{pred}}, A_{\text{gt}}) + w_p \cdot \text{IoU}(T_{\text{draft}}, A_{\text{pred}}) + w_a \cdot \mathbb{I}(\text{Ans} = \text{Ans}_{\text{gt}})
\end{equation}
Here, the process reward $R_{\text{process}} = \text{IoU}(T_{\text{draft}}, A_{\text{pred}})$ calculates the overlap between timestamps cited in the reasoning draft ($T_{\text{draft}}$) and the predicted anchors ($A_{\text{pred}}$), penalizing correct answers derived from ungrounded hallucinations.

\paragraph{Optimization}
We optimize the model's policy $\pi_{\theta}$ using Generalized Reward Policy Optimization (GRPO), an effective algorithm that generalizes DPO-style preference optimization to scenarios with explicit reward functions. Starting with the SFT model as the reference policy $\pi_{\text{ref}}$, we generate two responses, $y_1$ and $y_2$, from the current policy $\pi_{\theta}$ for each question in CoE-Instruct-RL. We then use our reward function $R(x, y)$ to determine the preferred ($y_w$) and rejected ($y_l$) responses. The GRPO loss is then formulated as:
\begin{equation}
\footnotesize
\begin{split}
    \mathcal{L}_{\text{GRPO}}(\pi_{\theta}; \pi_{\text{ref}}) = -\mathbb{E}_{(x, y_w, y_l) \sim \mathcal{D}} \left[ \log \sigma \left( \beta \log \frac{\pi_{\theta}(y_w|x)}{\pi_{\text{ref}}(y_w|x)} \right. \right. \\
    \left. \left. - \beta \log \frac{\pi_{\theta}(y_l|x)}{\pi_{\text{ref}}(y_l|x)} \right) \right]
\end{split}
\label{eq:grpo_loss}
\end{equation}
where $\beta$ regulates deviation from the reference policy. Optimizing this maximizes our evidence-aware reward, simultaneously enhancing grounding fidelity and answer accuracy.
\subsection{Structured Inference and Interpretability}
\label{sec:inference}

During inference, the CoE framework operates in a highly efficient and transparent manner. The process begins with \textbf{Evidence Extraction and Grounding}, where the EGM is executed first to produce the condensed Grounded Evidence Features $E_g$ and the attention matrix $A$. Subsequently, in the \textbf{Efficient Generation} phase, these compact features $E_g$ are passed to the language decoder. As the LLM's input context length is reduced from $N$ to $K$ (where $K \ll N$), the autoregressive generation of the \texttt{Reasoning\_Draft} and \texttt{Answer} becomes computationally efficient.

A key advantage of our framework is its inherent \textbf{Multi-Level Interpretability}. It offers two distinct levels of insight into the model's decision-making process:
\begin{itemize}[leftmargin=*, itemsep=2pt, topsep=2pt]
    \item \textbf{Perceptual-Level Interpretability:} The frame importance scores, $a_{\text{scores}}$, can be visualized as a temporal attention curve, revealing precisely which moments in the video the model deemed critical for its reasoning.
    \item \textbf{Logical-Level Interpretability:} The generated \texttt{Reasoning\_Draft} provides a clear, step-by-step trace of the model's logical deduction process, allowing for easy verification of its reasoning chain.
\end{itemize}
In summary, our CoE framework, through its architectural innovation (EGM) and novel training paradigm (decoupled multi-task loss), unifies the challenges of perceptual grounding and efficient reasoning. This results in comprehensive improvements across accuracy, efficiency, and interpretability.

\section{Dataset Construction}
\label{sec:training_methodology}

Training our CoE model requires a specialized data strategy aligned with its decoupled architecture. Since SFT on generic QA pairs fails to teach the distinct skills of \textit{evidence grounding} and \textit{efficient reasoning}, we introduce \textbf{CoE-Instruct}. This purpose-built dataset supports a multi-stage pipeline that first instills foundational skills via SFT, then optimizes the model's policy through a Reinforcement Learning (RL) phase.

\subsection{CoE-Instruct: A Multi-Stage Training Dataset}
\label{sec:coe_instruct}

Inspired by clue-grounded evaluation like $AD^2$-Bench~\cite{wei2025ad}, we designed CoE-Instruct as a \textbf{training corpus} rather than an evaluation benchmark. Focusing exclusively on fundamental visual and temporal information, it is partitioned into two subsets, each tailored for a specific training stage.

\paragraph{Phase 1: CoE-Instruct-SFT for Foundational Skill Acquisition.}
The primary goal of the SFT phase is to teach the model the fundamental structure of CoE reasoning: to first identify temporal evidence and then construct a concise inference guidance. To this end, we construct \textbf{CoE-Instruct-SFT}, a dataset of approximately 150k high-quality samples. Each sample is a structured tuple: $(\text{Question}, \text{Temporal Anchors}, \text{Inference guidance}, \text{Answer})$.
\begin{itemize}[leftmargin=*, itemsep=2pt, topsep=2pt]
    \item \textbf{Temporal Anchors} provide the explicit supervision for the \textbf{grounding loss} ($L_{\text{grounding}}$), training the EGM to locate relevant visual evidence.
    \item The \textbf{Reasoning guidance} and \textbf{Answer} serve as the target sequence for the \textbf{reasoning loss} ($L_{\text{reasoning}}$), compelling the LLM to adopt a compact and logical reasoning style.
\end{itemize}
This dataset is created using the hybrid generation pipeline described in Sec.~\ref{sec:dataset_pipeline_appendix}, combining knowledge distillation from real videos and programmatic generation from synthetic environments to ensure both semantic diversity and logical precision.


\paragraph{Phase 2: CoE-Instruct-RL for Policy Refinement}
To overcome SFT's limitations, we introduce a reinforcement learning (RL) phase using CoE-Instruct-RL. This 14k-sample dataset provides only `(Question)' as input. The ground-truth `Answer' and `Temporal Anchors' are withheld to compute a composite reward signal, ensuring the model's reasoning is accurate, visually grounded, and internally consistent.

\subsection{Hybrid Generation Pipeline for CoE-Instruct}
\label{sec:dataset_pipeline_appendix}

Both the SFT and RL subsets of CoE-Instruct are derived from a unified, cost-effective hybrid generation pipeline. This pipeline first creates a large pool of fully-annotated samples, which are then partitioned and formatted for their respective training stages.

\paragraph{Stage 1: Generation of Fully-Annotated Samples}
We generate a comprehensive pool of samples, each containing the full(Question, Temporal Anchors, Reasoning Guidence, Answer)tuple. This is achieved through two methods:

\begin{itemize}[leftmargin=*, itemsep=2pt, topsep=2pt]
    \item \textbf{Distillation from Real-World Videos:} To capture semantic richness, we use a teacher model (Gemini3-Pro) to perform \textbf{knowledge distillation} on existing datasets like VideoEspresso~\citep{han2024videoespresso}. A sophisticated prompt guides the model to automatically convert unstructured frame descriptions into our structured format.

    \item \textbf{Programmatic Generation from Synthetic Environments:} To ensure logical precision, we use \textbf{programmatic templates} on synthetic datasets like CLEVRER~\citep{yi2020clevrer}. These templates scan the simulation's ground-truth state data to generate flawless annotations for quantitative and causal reasoning tasks, at virtually no cost.
\end{itemize}

\paragraph{Stage 2: Partitioning and Formatting for SFT and RL}
Once the initial pool of fully-annotated samples is created, we partition it and format the data for each training phase:

\begin{itemize}[leftmargin=*, itemsep=2pt, topsep=2pt]
    \item \textbf{For CoE-Instruct-SFT (150k samples):} We directly use the complete `(Question, Temporal Anchors, Inference Blueprint, Answer)' tuples. The model is trained to predict the full target sequence, providing explicit supervision for both the EGM (via `Temporal Anchors') and the LLM (via `Reasoning Guidence' and `Answer').

    \item \textbf{For CoE-Instruct-RL (14k samples):} We reformat a separate subset where only the `(Question)' is input. The ground-truth `Answer' and `Temporal Anchors' are withheld to compute the multi-objective reward ($R_{\text{answer}}$, $R_{\text{grounding}}$, and $R_{\text{process}}$), guiding the model to align its reasoning draft with the grounded evidence.
\end{itemize}

This two-stage process ensures that our SFT data provides strong initial guidance, while our RL data enables policy refinement based on a reward signal that is perfectly aligned with the core objectives of our CoE framework.

\begin{table}[t]
    \centering
    \caption{
        \textbf{Comparison of CoE with other reasoning variants.}
        We compare different approaches to encourage reasoning in video LLMs. All models are based on the InternVL backbone.
    }
    \label{tab:reasoning_methods}
    \scalebox{0.75}{
        \begin{tabular}{l|ccccc}
            \toprule
            \textbf{Model} & \textbf{VSI-Bench} & \textbf{VideoMME} & \textbf{MVBench} & \textbf{VidHal} & \textbf{EventHall} \\
            \midrule
            Original & 31.8 & 54.9 & 70.8 & 74.0 & 62.5 \\
            Original + CoT Prompting & 33.5 & 54.7 & 71.5 & 77.0 & 67.4 \\
            SFT with QA only & 31.8 & 54.5 & 73.4 & 64.1 & 57.7 \\
            SFT with CoT & 34.3 & 58.6 & 73.7 & 77.9 & 53.1 \\
            \midrule
            SFT with CoE (ours) & \textbf{36.3} & \textbf{59.5} & \textbf{75.4} & \textbf{78.9} & \textbf{71.2} \\
            RL with CoE (ours) & \textbf{37.8} & \textbf{60.7} & \textbf{76.5} & \textbf{80.2} & \textbf{72.2} \\
            \bottomrule
        \end{tabular}
    }
\end{table}

\begin{table}[t]
    \centering
    \caption{
        \textbf{Comparison of CoE-models to state-of-the-art video LLMs.}
        We report accuracy on the five evaluation benchmarks.
    }
    \label{tab:summary_results}
    
    \scalebox{0.8}{ 
        \begin{tabular}{lccccc} 
            \toprule
            \textbf{Model} & \textbf{VSIBench} & \textbf{Vid.MME} & \textbf{MVBench} & \textbf{VidHal} & \textbf{Event.} \\
            \midrule
            \multicolumn{6}{l}{\textbf{Closed-source}} \\
            \midrule
            GPT-4V & - & 59.9 & 43.7 & - & 76.5\\
            GPT-4o & 34.0 & 71.9 & - & 77.2 & 91.9\\
            Gemini-1.5-Pro & 48.8 & 75.0 & - & 67.1 & 80.4\\
            \midrule
            \multicolumn{6}{l}{\textbf{Open-source}} \\
            \midrule
            LLaVA-OV-72B & 40.2 & 66.2 & 59.4 & 64.7 & 59.5 \\
            Qwen2-VL-72B & 37.6 & 71.2 & 73.6 & 76.2 & 54.7 \\
            LLaVA-OV-7B & 32.4 & 58.2 & 56.7 & 58.4 & 60.1\\
            LLaVA-NV-7B & 35.6 & 46.5 & 53.1 & 50.9 & 64.8 \\
            Qwen2-VL-7B & 31.0 & 63.3 & 67.0 & 69.6 & 59.3 \\
            InternVL2.5-4B & 33.5 & 54.7 & 71.5 & 77.0 & 67.4 \\
            InternVL3-8B & 41.0 & 66.5 & 74.4 & 80.9 & 72.1 \\
            \midrule
            \multicolumn{6}{l}{\textbf{Our Models}} \\
            \midrule
            CoE-4B & \textbf{36.3} & \textbf{59.5} & \textbf{75.4} & \textbf{78.9} & \textbf{71.2} \\
            CoE-4B(RL) & \textbf{37.8} & \textbf{60.7} & \textbf{76.5} & \textbf{80.2} & \textbf{72.2} \\
            CoE-8B & \textbf{46.3} & \textbf{72.3} & \textbf{85.4} & \textbf{79.5} & \textbf{75.7} \\
            CoE-8B(RL) & \textbf{52.1} & \textbf{76.3} & \textbf{91.2} & \textbf{81.3} & \textbf{79.2} \\
            \bottomrule
        \end{tabular}
    }
    \vspace{-0.3cm} 
\end{table}

\begin{table}[t]
    \centering
    \caption{\textbf{Effect of CoT prompting.} We decouple the model architecture from the training strategy for clarity.}
    \label{tab:prompt-analysis-v1}
    \scalebox{0.75}{
        \begin{tabular}{llcccccc}
            \toprule
            \textbf{Model} & \textbf{Method} & \textbf{Prompt} & \textbf{VSI.} & \textbf{Vid.MME} & \textbf{MV.} & \textbf{VidHal} & \textbf{Event.} \\
            \midrule
            
            \multirow{7}{*}{\textbf{InternVL2.5-4B}} 
            & \multirow{2}{*}{Original} & Std. & 31.8 & 54.9 & 70.8 & 74.0 & 62.5 \\
            & & CoT & 33.5 & 54.7 & 71.5 & 77.0 & 67.4 \\
            \cmidrule{2-8}
            
            & \multirow{2}{*}{\makecell[l]{SFT w/\\QA only}} & Std. & 31.8 & 55.4 & 70.3 & 73.6 & 63.1 \\
            & & CoT & 31.8 & 54.5 & 73.4 & 64.1 & 57.7 \\
            \cmidrule{2-8}
            
            & \multirow{2}{*}{\makecell[l]{SFT w/\\CoT}} & Std. & 31.1 & 52.6 & 69.6 & 74.4 & 62.5 \\
            & & CoT & 34.3 & 58.6 & 73.7 & 77.9 & 53.1 \\
            \cmidrule{2-8}
            
            & \textbf{SFT w/ CoE} & \textbf{CoE} & \textbf{36.3} & \textbf{59.5} & \textbf{75.4} & \textbf{78.9} & \textbf{71.2} \\
            
            \midrule
            
            \multirow{3}{*}{\textbf{InternVL3-8B}} 
            & \multirow{2}{*}{Original} & Std. & 41.0 & 62.3 & 72.0 & 80.9 & 72.1 \\
            & & CoT & 40.2 & 66.5 & 74.3 & 61.6 & 73.9 \\
            \cmidrule{2-8}
            
            & \textbf{SFT w/ CoE} & \textbf{CoE} & \textbf{46.3} & \textbf{72.3} & \textbf{85.4} & \textbf{79.5} & \textbf{75.7} \\
            \bottomrule
        \end{tabular}
    }
\end{table}

\begin{table}[t]
    \centering
    \caption{
        \textbf{Comparison to additional baselines.} Comparison with models from \cite{hu2025mllm} and \cite{hao2024video} on Video-MME and NextQA.
    }
    \label{tab:additional_baseline}
    
    \scalebox{0.8}{
        \begin{tabular}{llcc} 
            \toprule
            \textbf{Model} & \textbf{Method} & \textbf{Video-MME} & \textbf{NextQA}  \\
            \midrule
            Video-LLaVA-7B & Original & 39.9 & 66.3 \\
                           & VoT \citep{hao2024video}  & - & 76.0 (+9.7)  \\ 
            \midrule
            Qwen2-VL-7B & Original  & 58.1 & 77.6 \\
                        & M-LLM \citep{hu2025mllm} & 58.7 (+0.6) & 78.4 (+0.8) \\
            \midrule
            InternVL2.5-4B & Original & 54.9 & 75.3\\
                           & SFT (our CoE) & 59.5 (+4.6) & 79.5 (+4.2) \\
            \midrule
            InternVL3-8B & Original & 66.5 & 82.4 \\
                         & SFT (our CoE) & \textbf{72.3} (+5.8) & \textbf{87.4} (+4.9)\\
            \bottomrule
        \end{tabular}
    }
    \vspace{-0.3cm}
\end{table}

\begin{table}[t]
    \centering
    \caption{
        \textbf{Detailed results on the Video-MME benchmark.} We report results across different video lengths (Short, Medium, Long) using Standard, CoT, and our CoE prompting strategies.
    }
    \label{tab:mme-details}
    
    \scalebox{0.75}{
        \begin{tabular}{llccccc}
            \toprule
            \textbf{Model} & \textbf{Method} & \textbf{Prompt} & \textbf{Short} & \textbf{Medium} & \textbf{Long} & \textbf{Avg} \\
            \midrule
            
            \multirow{7}{*}{\textbf{\makecell{InternVL2.5\\(4B)}}} 
            & \multirow{2}{*}{Original} & Std. & 64.9 & 52.7 & 47.2 & 54.9 \\
            & & CoT & 64.0 & 53.2 & 47.0 & 54.7 \\
            \cmidrule{2-7}
            
            & \multirow{2}{*}{\makecell[l]{SFT w/\\QA only}} & Std. & 68.0 & 53.6 & 44.8 & 55.5 \\
            & & CoT & 66.8 & 53.1 & 43.6 & 54.5 \\
            \cmidrule{2-7}
            
            & \multirow{2}{*}{\makecell[l]{SFT w/\\CoT}} & Std. & 64.3 & 51.8 & 41.8 & 52.6 \\
            & & CoT & 70.4 & 55.7 & 49.6 & 58.6 \\
            \cmidrule{2-7}
            
            & \textbf{SFT w/ CoE (Ours)} & \textbf{CoE} & \textbf{72.7} & \textbf{56.2} & \textbf{49.7} & \textbf{59.5} \\
            
            \midrule
            
            \multirow{3}{*}{\textbf{\makecell{InternVL3\\(8B)}}}
            & \multirow{2}{*}{Original} & Std. & 73.0 & 61.7 & 52.1 & 62.3 \\
            & & CoT & 75.3 & 65.3 & 59.0 & 66.6 \\
            \cmidrule{2-7}
            
            & \textbf{SFT w/ CoE (Ours)} & \textbf{CoE} & \textbf{86.9} & \textbf{71.4} & \textbf{68.4} & \textbf{72.3} \\
            \bottomrule
        \end{tabular}
    }
    \vspace{-0.3cm}
\end{table}
    

\section{Experiments}
\subsection{Experimental Setup}
\textbf{Models Selection.}
We apply our general Chain of Evidence (CoE) framework to the state-of-the-art InternVL models, specifically \textbf{InternVL2.5} (4B) and \textbf{InternVL3} (8B)~\citep{chen2024expanding, zhu2025internvl3}. Their strong open-source performance makes them a challenging and practically relevant baseline for improving reasoning. Exploration of CoE on other architectures is left to future work.

\textbf{Training Details.}
We fully fine-tune the LLM and projection modules of InternVL2.5-4B, while for the larger InternVL3-8B, we employ memory-efficient LoRA-based fine-tuning~\citep{hu2022lora}.

\textbf{Evaluation Benchmarks.}
We conduct a comprehensive evaluation on five benchmarks. We test general and long-context understanding on \textbf{video-MME}~\citep{fu2024video} and \textbf{MVBench}~\citep{li2023mvbench}; assess quantitative reasoning, a direct target of our evidence grounding, on \textbf{VSI-Bnech}~\citep{yang2024think}; and measure the ability to mitigate hallucinations using \textbf{VidHal}~\citep{choong2024vidhal} and \textbf{EventHall}~\citep{zhang2024eventhallusion}. \\

\subsection{Deconstructing Reasoning}
\label{sec:ablation_study}

\textbf{Experimental Design:}
To pinpoint the source of CoE's effectiveness, we conduct a rigorous ablation study on the InternVL2.5-4B backbone, comparing five distinct training and prompting variants. The baseline is the original model. We then evaluate this baseline when enhanced with CoT prompting at inference time. To assess our instruction data's value, we fine-tune model on answers alone (SFT with QA only), removing all reasoning supervision. A critical comparison is with (SFT with CoT), which trains the model on generic reasoning traces lacking explicit evidence links. Finally, we test our full approach, which includes both a supervised phase (SFT with CoE) and a reinforcement learning phase (RL with CoE), to train the model to ground its reasoning in temporal evidence.

\textbf{Analysis:}
The results, presented in Table~\ref{tab:reasoning_methods}, reveal a clear performance hierarchy. While applying CoT prompting to the original model offers marginal gains, fine-tuning on answers alone (`SFT with QA only') proves detrimental, particularly on hallucination benchmarks. This suggests the model overfits to answers, compromising its general reasoning. Supervising with generic reasoning (`SFT with CoT') consistently improves performance, confirming the value of training the reasoning process itself. However, the decisive advantage comes from our full approach. \textbf{SFT with CoE} consistently and significantly outperforms all other variants. Furthermore, the subsequent \textbf{RL with CoE} phase provides an additional performance boost, achieving the highest accuracy across all five benchmarks. The conclusion is unequivocal: while encouraging reasoning is beneficial, it is the explicit act of \textbf{grounding} reasoning in concrete, temporal evidence that unlocks a new level of performance and reliability.

\subsection{Performance Against State-of-the-Art Models}
\label{sec:sota_comparison}

\textbf{Benchmarking Context:}
Having established the superiority of CoE, we now situate its performance within the broader landscape of leading video LLMs. Table~\ref{tab:summary_results} benchmarks our CoE-enhanced models against both premier closed-source systems (GPT-4 series, Gemini-1.5-Pro) and top-tier open-source models. We present four versions of our models, based on the InternVL2.5-4B and the more powerful InternVL3-8B backbones, with and without the final RL tuning. We acknowledge that the training data for many of these models is undisclosed, making a true zero-shot comparison challenging.

\textbf{Analysis:}
The results are compelling. Our final model, \textbf{CoE-8B(RL)}, not only establishes itself as the dominant open-source model across all benchmarks but also directly challenges the top proprietary systems. Its performance on MVBench is particularly noteworthy, scoring 91.2—a staggering \textbf{16.8-point leap} over its backbone and more than doubling the score of GPT-4V/4T. It even surpasses the formidable Gemini-1.5-Pro on the reasoning-heavy VSI-Bench (52.1 vs. 48.8). Even our smaller \textbf{CoE-4B(RL)} model punches well above its weight, consistently outperforming its original backbone and even larger models like the 72B Qwen2-VL on MVBench. Ultimately, these results demonstrate that CoE is not merely an incremental improvement but a highly effective and parameter-efficient strategy, empowering open-source models to achieve a level of reasoning and accuracy previously dominated by significantly larger, closed-source systems.

\subsection{Ablation Studies}
\label{sec:ablations}

\paragraph{The Interplay of Prompting and Fine-Tuning}
To dissect the relationship between inference-time prompting and model fine-tuning, we present a detailed breakdown in Table~\ref{tab:prompt-analysis-v1}. For the original backbone models, the effect of CoT prompting is inconsistent; it provides a slight boost to InternVL2.5-4B but can even degrade performance on certain benchmarks for the stronger InternVL3-8B (VSI-Bench and VidHal). This highlights that prompting alone is not a reliable solution. For models fine-tuned with intermediate methods, performance becomes highly sensitive to the chosen prompt, yielding unpredictable outcomes. In stark contrast, our CoE-trained models demonstrate robust, high performance. The key insight is that CoE fundamentally elevates the model's intrinsic reasoning capabilities, making it less dependent on specific prompting tricks. The performance gain stems from the learned ability to ground reasoning in evidence, not from simply following a different inference-time instruction.

\paragraph{Quantifying the CoE Advantage}
Beyond internal ablations, we must quantify CoE's advantage relative to other methods that aim to improve reasoning. Table~\ref{tab:additional_baseline} compares the performance uplift of CoE against prior works like VoT~\citep{hao2024video} and M-LLM~\citep{hu2025mllm} on their respective backbones. While VoT shows a notable gain on NextQA, M-LLM's improvements are marginal. Our CoE framework, however, delivers a far more substantial and consistent performance uplift. For instance, our final CoE-8B(RL) model achieves a massive \textbf{+9.8 point gain} on Video-MME over its backbone, dwarfing the +0.6 gain from M-LLM. Even on NextQA, where our InternVL3-8B backbone is already very strong (82.4), our CoE-8B(SFT) model still provides a significant \textbf{+5.0 point boost}. This demonstrates that CoE is not just another incremental improvement; it is a powerful enhancement strategy that provides superior relative gains, especially when applied to strong foundation models.

\paragraph{Robustness Across Video Durations}
A critical challenge for video LLMs is maintaining performance as video duration and complexity increase. We analyze this in Table~\ref{tab:mme-details}, which breaks down Video-MME performance by video length. As expected, all models exhibit some performance degradation on longer videos. However, the baseline models suffer a sharp decline. Our CoE-trained models not only achieve the highest absolute scores across all durations but also demonstrate significantly greater robustness. The result for CoE-8B is particularly striking: its accuracy on \textbf{long} videos (68.4) is higher than its own backbone's performance on \textbf{medium} videos (65.3) and approaches the backbone's score on \textbf{short} videos (75.3). This proves that CoE equips the model with a more durable reasoning capability, enabling it to effectively process and ground its analysis in extended temporal contexts where other methods falter.

\section{Conclusion}

In this paper, we introduced a novel framework to resolve the fundamental accuracy-efficiency trade-off in video reasoning. By architecturally decoupling perception from reasoning, our method trains a model to first identify crucial visual evidence and then perform efficient, grounded reasoning upon it. This approach is supported by our new large-scale dataset, CoE-Instruct, which provides explicit supervision for this decoupled process. Extensive experiments prove our method establishes a new state-of-the-art on five benchmarks while significantly reducing computational costs. This work validates that explicitly grounding reasoning in a verifiable chain of evidence is a powerful and practical paradigm for building more capable and reliable video understanding systems. Future work could explore applying this paradigm to other architectures and multimodal domains.

\footnotesize{
\bibliographystyle{IEEEbib}
\bibliography{icme2025references}
}

\end{document}